# A Geometric Method to Obtain the Generation Probability of a Sentence

Chen Lijiang
Nanjing Normal University
ljchen97@126.com

**Abstract**   "How to generate a sentence" is the most critical and difficult problem in all the natural language processing technologies. In this paper, we present a new approach to explain the generation process of a sentence from the perspective of mathematics. Our method is based on the premise that in our brain a sentence is a part of a word network which is formed by many word nodes. Experiments show that the probability of the entire sentence can be obtained by the probabilities of single words and the probabilities of the co-occurrence of word pairs, which indicate that human use the synthesis method to generate a sentence.

How the human brain realizes natural language understanding is still an open question. With the development of statistical natural language processing technology, people gradually realized that we should use the possibility to determine whether a sentence is reasonable, which means use the probability to measure the grammatical well-formedness of a sentence. If the probability of the first sentence is $10^{-15}$, the probability of the second sentence is $10^{-18}$, and the probability of the third sentence is $10^{-20}$. Then the first sentence is most likely correct sentence, which is 1000 times more likely to be correct than the second sentence, and 100,000 times than the third sentence.

We use $P(x_1 x_2 \cdots x_{n-1} x_n)$ to represent the probability of the sentence, where n is the number of words in a sentence, $x_i$ and $x_j$ $(1 \leq i, j \leq n)$ are two words in a sentence. Using the conditional probability formula, $P(x_1 x_2 \cdots x_{n-1} x_n)$ can be expanded as follows:

$$P(x_1 x_2 \cdots x_{n-1} x_n) = P(x_1) \bullet P(x_2 | x_1) \bullet P(x_3 | x_1, x_2) \cdots P(x_n | x_1, x_2, \cdots x_{n-1})$$

If we have enough data, the probability of the first word $P(x_1)$ and the conditional probability of the second word $P(x_2 | x_1)$ should be very easy to calculate. However, because of fewer occurrences of $x_1 x_2 x_3$ in the corpus, it is very difficult to calculate the conditional probability of the third word $P(x_3 | x_1, x_2)$. When n is large enough, even we have enough data, the number of times that $x_1 x_2 \cdots x_{n-1} x_n$ appears may be very small, and $x_1 x_2 \cdots x_{n-1} x_n$ even never appears in the data files. Thus estimating $P(x_1 x_2 \cdots x_{n-1} x_n)$ becomes an impossible task. We call this **Data Sparsity Problem**.

In order to overcome the difficulty of data sparsity, a famous mathematician Andrey Markov found a solution that assumes the probability of any word is only related to the words in front of it. Thus find the probability of such a sentence becomes simple:

$$P(x_1 x_2 \cdots x_{n-1} x_n) = P(x_1) \bullet P(x_2 | x_1) \bullet P(x_3 | x_2) \cdots P(x_n | x_{n-1})$$

But this method would ignore many long-distance relationships between words in a sentence, such as $P(x_3 | x_1)$、$P(x_5 | x_2)$ et al. Therefore, to determine the exact probability of a sentence, it is necessary to find a new approach.

We use a network structure of words to show the relationship between words (Figure 1), in which each dot represents a word and each edge represents a relationship between two words. The edge between two dots indicates that there is an association between these two words; otherwise, they are independent of each other. Weights can be added to the edges, for instance, the probability of simultaneous occurrence of two words. The dashed line in Figure 1 represents the area that contains all the words to form a complete sentence. We found Figure 1 is similar to the brain neural network. If each neuron in the neural network stands for a word (or other linguistic unit), the identified area in the figure can be viewed as an activated sentence.

**Generating a sentence can be seen as all the words in the sentence (or other linguistic units) are activated simultaneously in the neural network.** If the probability of a sentence can be obtained by the probability of each activated word (or other linguistic units) being activated, and the coincidence probability of an activated

word pair(or other linguistic units), as long as the language data is sufficient, we can find the probability of a sentence being generated under any language environment.

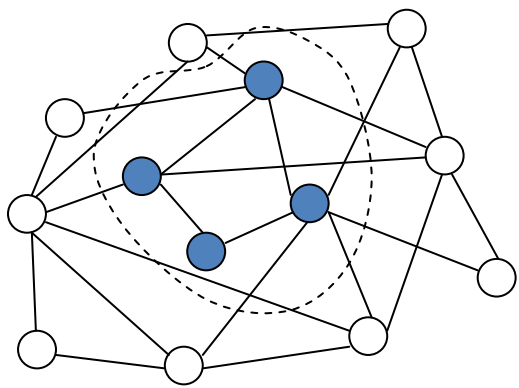

Figure 1    activated words in the network to form a sentence

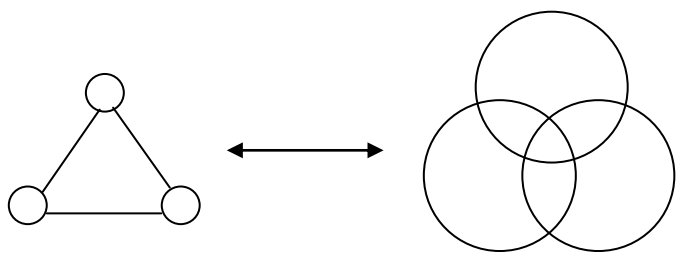

Figure 2    the relationship of three words can be represented as a network or a set diagram

In the probability theory, we have such two addition equations:
$$P(AB) = P(A \cup B) - P(A) - P(B) \qquad (1)$$

$$P(ABC) = P(A \cup B \cup C) - P(A) - P(B) - P(C) + P(AB) + P(BC) + P(AC) \quad (2)$$

Formula (2) tells us that if we want to know $P(ABC)$, not only need to know $P(A), P(B), P(C)$ and $P(AB), P(AC), P(BC)$, but also need to know $P(A \cup B \cup C)$.

Now, we have a way to obtain $P(ABC)$ under the premise that we only know $P(A), P(B), P(C)$ and $P(AB), P(AC), P(BC)$. **In the Venn diagram of three events (see Figure 3), an intuitive feeling is that there is a link between the center** **portion of the figure and** $P(ABC)$. By Venn diagram, we hope to derive the joint probability of three events, if the probability of one event $P(A), P(B), P(C)$, and the joint probability of two events $P(AB), P(AC), P(BC)$ are known. Here we experimentally validate our intuition.

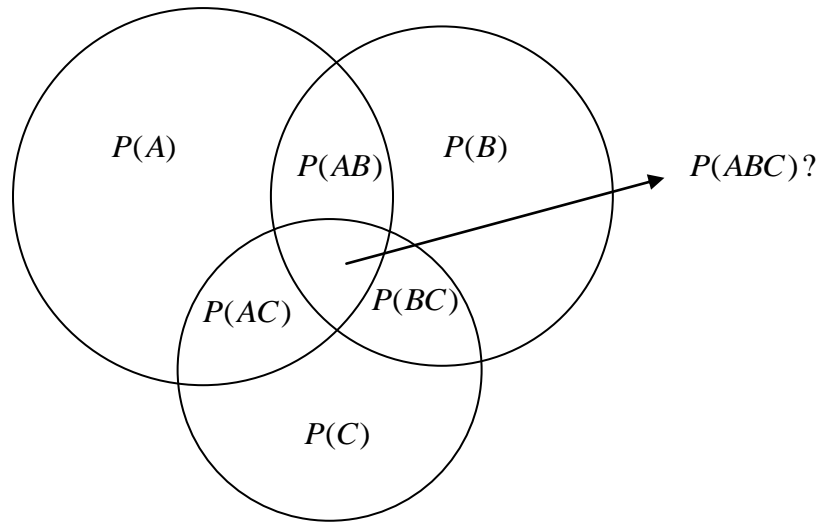

Figure 3 Whether the central part of the Venn diagram can represent $P(ABC)$

In order to test our hypothesis, we draw a Venn diagram under the premise that we know the $P(A), P(B), P(C)$ and $P(AB), P(AC), P(BC)$. Let the areas of three circles are equal to $P(A), P(B), P(C)$, then according to $P(AB), P(AC)$ and $P(BC)$ adjust the position of the three circles, so that each intersection area of two circles are equal to $P(AB), P(AC), P(BC)$. The detail is as the followings.

## 1 Calculating the Area of Center part of Venn diagram

We construct a Venn diagram representing the relationship between three events based on the known probabilities $P(A), P(B), P(C)$ and $P(AB), P(AC), P(BC)$. Firstly we resize the circle radiuses so the area of circles are equal to $P(A), P(B), P(C)$, and then adjust the distance between the two circle centers that $P(AB), P(AC), P(BC)$ are equal to intersection areas.

Step1: Let A, B, C respectively be the center of the three circles (See Figure 4). Set up $L_{AB} = r$、$L_{BC} = t$、$L_{AC} = s$.

By the calculation methods of plane geometry, we can get:

$$P(AB) = \pi a^2 \frac{2\arccos \frac{a^2 + r^2 - b^2}{2ar}}{2\pi} + \pi b^2 \frac{2\arccos \frac{b^2 + r^2 - a^2}{2br}}{2\pi}$$
$$- \frac{1}{2}\sqrt{(a+b+r)(a+b-r)(a+r-b)(b+r-a)}$$

$$P(BC) = \pi a^2 \frac{2\arccos \frac{b^2 + t^2 - c^2}{2bt}}{2\pi} + \pi b^2 \frac{2\arccos \frac{c^2 + t^2 - b^2}{2ct}}{2\pi}$$
$$- \frac{1}{2}\sqrt{(b+c+t)(b+c-t)(b+t-c)(c+t-b)}$$

$$P(AC) = \pi a^2 \frac{2\arccos \frac{a^2 + s^2 - c^2}{2as}}{2\pi} + \pi b^2 \frac{2\arccos \frac{c^2 + s^2 - a^2}{2cs}}{2\pi}$$
$$- \frac{1}{2}\sqrt{(a+c+s)(a+c-s)(a+s-c)(c+s-a)}$$

Among them $\pi a^2 = P(A)$、$\pi b^2 = P(B)$、$\pi c^2 = P(C)$

Based on the three above equations, we cannot get precise results of $r, s, t$. But we can use the split-half method to obtain the approximating solutions of these equations.

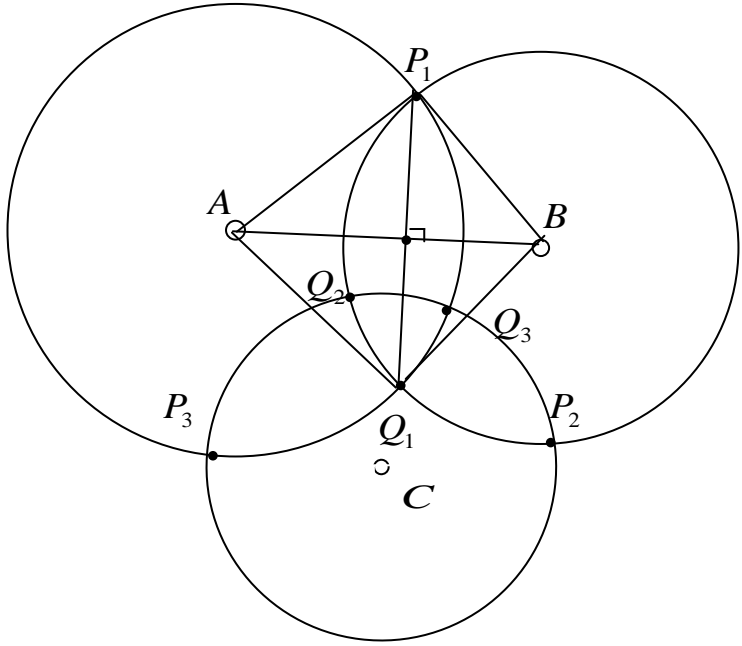

Figure 4 Calculate the distance between the two circle centers

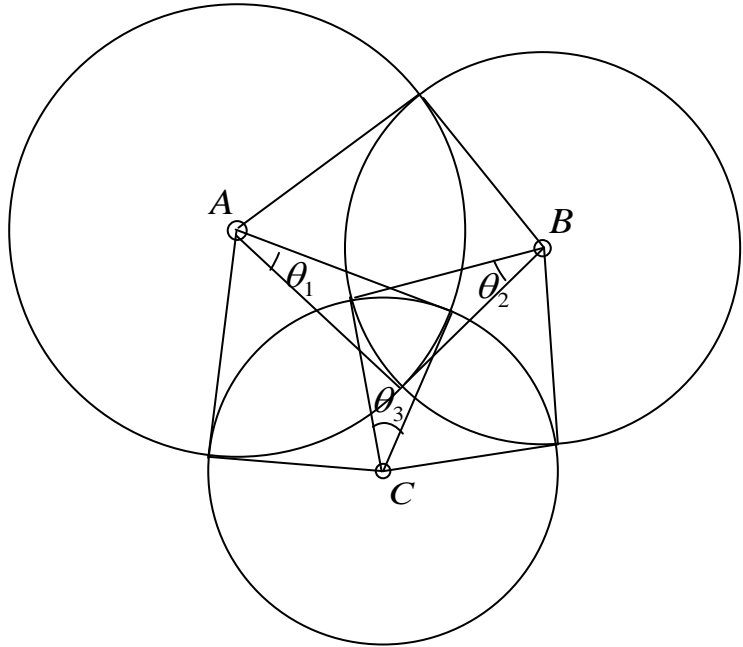

Figure 5 calculates the angle (1)

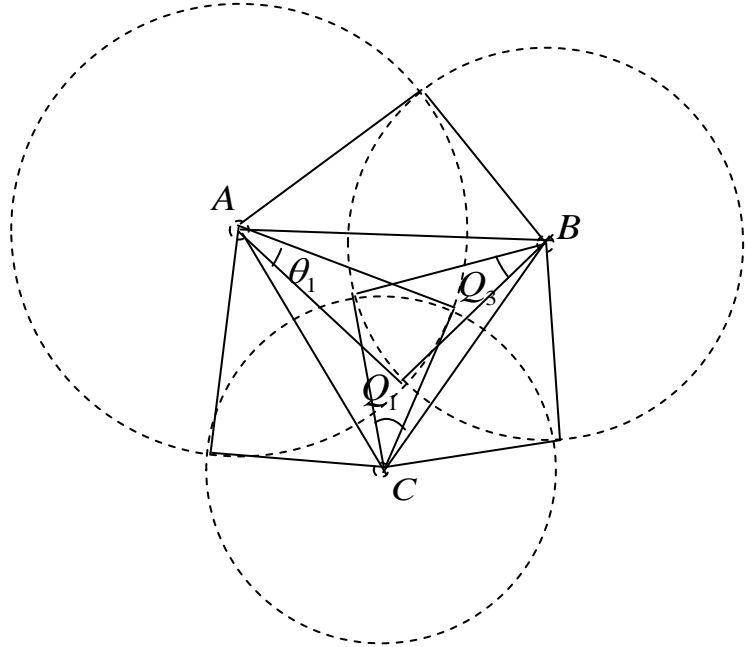

Figure 6 calculates the angle (2)

Step2: calculate $\theta_1$、$\theta_2$、$\theta_3$ (see Figure 5)

Known $L_{AB} = r$、$L_{BC} = t$、$L_{AC} = s$, $L_{AQ_1} = L_{AQ_3} = a$、$L_{BQ_1} = b$、$L_{CQ_3} = c$, we can determine the angle degree of $\triangle ABC$、$\triangle ABQ_1$、$\triangle ACQ_3$ according to the formulas below.

$\because \cos \angle CAQ_3 = \dfrac{a^2 + s^2 - c^2}{2as}$

$\cos \angle BAQ_1 = \dfrac{a^2 + r^2 - b^2}{2ar}$

$\cos \angle BAC = \dfrac{r^2 + s^2 - t^2}{2rs}$

$$\theta_1 = \angle BAC - \angle BAQ_3 - \angle CAQ_1$$
$$= \angle BAC - (\angle BAC - \angle CAQ_3) - (\angle BAC - \angle BAQ_1)$$
$$= \angle CAQ_3 + \angle BAQ_1 - \angle BAC$$

$$\therefore \theta_1 = \arccos \frac{a^2 + s^2 - c^2}{2as} + \arccos \frac{a^2 + r^2 - b^2}{2ar} - \arccos \frac{r^2 + s^2 - t^2}{2rs}$$

Similarly

$$\theta_2 = \arccos \frac{b^2 + t^2 - c^2}{2bt} + \arccos \frac{b^2 + r^2 - a^2}{2br} - \arccos \frac{r^2 + t^2 - s^2}{2rt}$$

$$\theta_3 = \arccos \frac{c^2 + s^2 - a^2}{2cs} + \arccos \frac{c^2 + t^2 - b^2}{2ct} - \arccos \frac{t^2 + s^2 - r^2}{2ts}$$

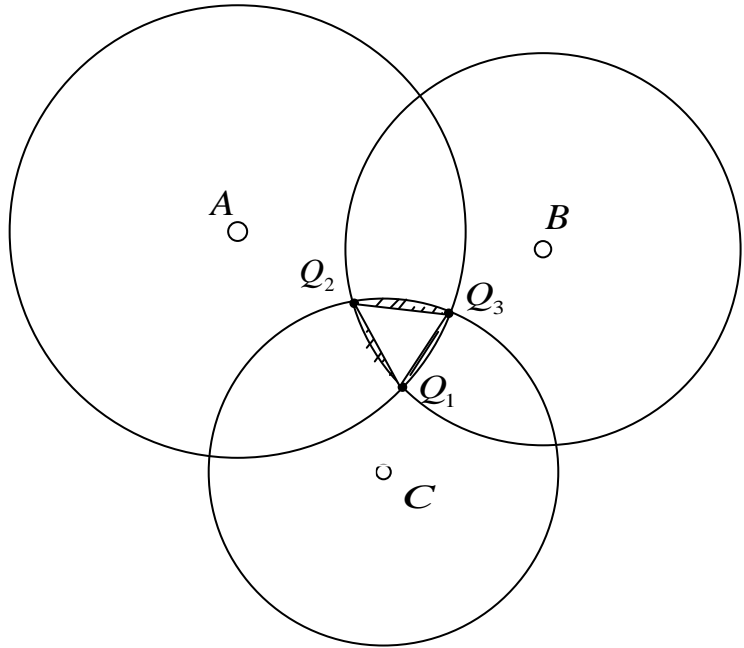

Figure 7　Calculate the area of the central part

Step 3: determine the areas of three figures bounded by arc $Q_1Q_2$ and line $Q_1Q_2$ ($S_1$), arc $Q_1Q_3$ and line $Q_1Q_3$ ($S_2$), arc $Q_2Q_3$ and line $Q_2Q_3$ ($S_3$) (Figure 7)

$$S_1 = \frac{\theta_1}{360°} \times \pi a^2 - \sin\frac{\theta_1}{2} \cos\frac{\theta_1}{2} a^2$$

$$S_2 = \frac{\theta_2}{360°} \times \pi b^2 - \sin\frac{\theta_2}{2} \cos\frac{\theta_2}{2} b^2$$

$$S_3 = \frac{\theta_3}{360°} \times \pi c^2 - \sin\frac{\theta_3}{2}\cos\frac{\theta_3}{2}c^2$$

Step 4: calculate he area of triangle $\Delta Q_1 Q_2 Q_3$.

$$L_{Q_1 Q_3} = 2a\sin\frac{\theta_1}{2}$$

$$L_{Q_1 Q_2} = 2b\sin\frac{\theta_2}{2}$$

$$L_{Q_2 Q_3} = 2c\sin\frac{\theta_3}{2}$$

Acoording to Helen formula

$$S_{\Delta Q_1 Q_2 Q_3} = \frac{1}{4}\sqrt{(2a\sin\frac{\theta_1}{2} + 2b\sin\frac{\theta_2}{2} - 2c\sin\frac{\theta_3}{2})(2a\sin\frac{\theta_1}{2} + 2b\sin\frac{\theta_2}{2} - 2c\sin\frac{\theta_3}{2})}$$

$$\overline{(2a\sin\frac{\theta_1}{2} + 2c\sin\frac{\theta_3}{2} - 2b\sin\frac{\theta_2}{2})(2b\sin\frac{\theta_2}{2} + 2c\sin\frac{\theta_3}{2} - 2a\sin\frac{\theta_1}{2})}$$

Step 5: $S_{cantral\ part} = S_1 + S_2 + S_3 + S_{\Delta Q_1 Q_2 Q_3}$

## 2 Experiment Analysis

We simulate the random trial by generating random numbers. Let event A be "word A appears in a sentence", event B be "word B appears in a sentence", and event C be "word C appears in a sentence". To

calculate $P(A), P(B), P(C), P(AB), P(AC), P(BC)$ and $P(ABC)$, we must generate

enough test results randomly. Each result is simplified as a two-value triple, such as

(1,0,1) which means Word A appears in the sentence, word B does not appear in the sentence, and the word C appears in the sentence(Table 1 shows 5 examples).

| n | A | B | C |
|---|---|---|---|
| 1 | 0 | 1 | 1 |
| 2 | 1 | 0 | 1 |
| 3 | 0 | 0 | 1 |
| 4 | 1 | 1 | 0 |
| 5 | 1 | 0 | 0 |

Table 1  5 results of the random trial

Firstly, we randomly generated 32,000 copies of probabilistic data. Each copy is obtained based on 10,000 independent test results. We can get $P(A), P(B), P(C), P(AB), P(AC), P(BC)$ and $P(ABC)$ by analyzing 10,000 results statistically, and calculate the area S of the central part of Venn diagram. Then all the data are sorted based on the size of $P(ABC)$. The diagram below shows the relationship between the area S of the center portion and $P(ABC)$:

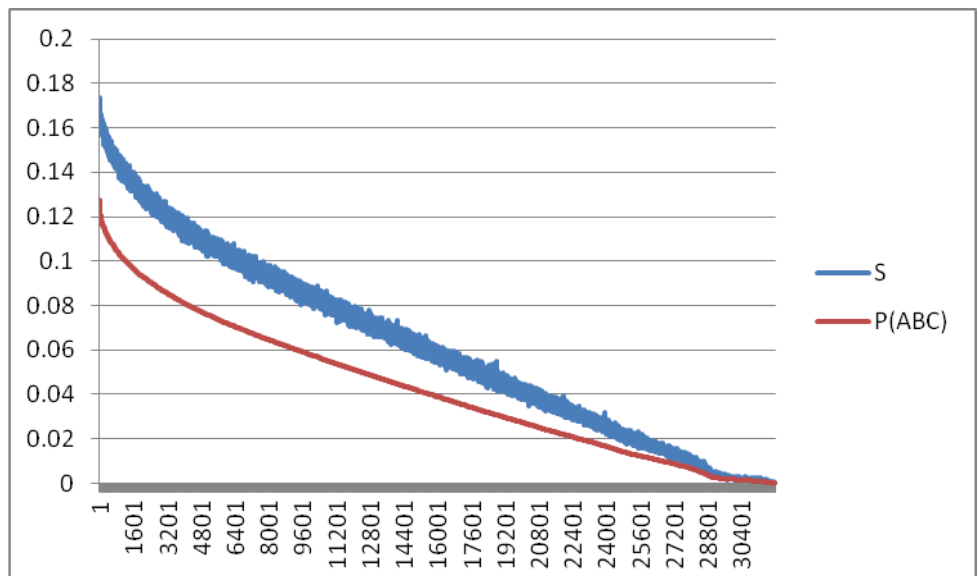

Figure 8 when the test number of is 10,000, the curves of the area S of the central part of Venn diagram and $P(ABC)$

Then, we randomly generated 32,000 copies of probabilistic data, and each copy is obtained based on 1,000,000 independent test results. The diagram below shows the relationship between the area S of the center portion and $P(ABC)$:

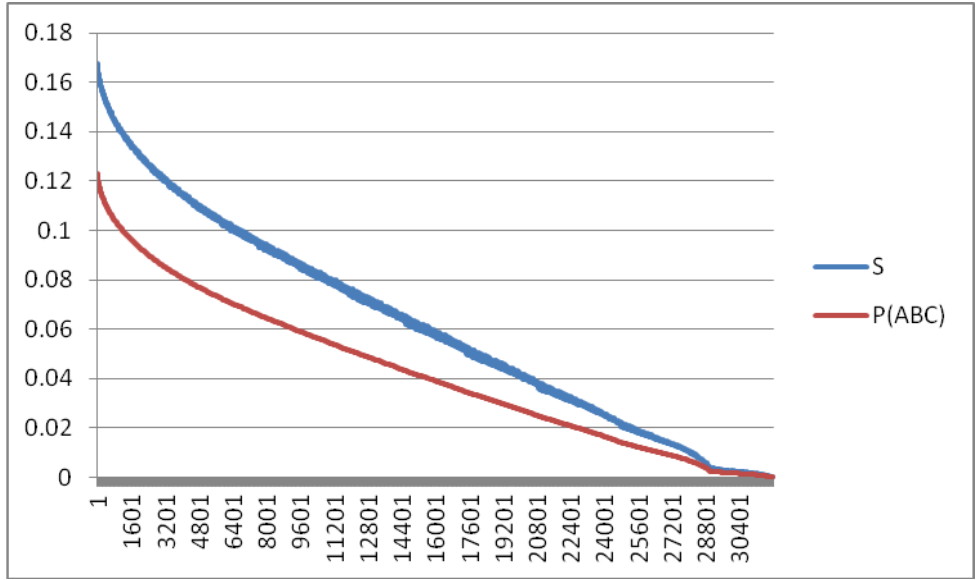

Figure 9 when the test number of is 1,000,000, the curves of the area S of the central part of Venn diagram and $P(ABC)$

Finally, the number of tests is increased to 100,000,000 per copy. The diagram below shows the relationship between the area S of the center portion and $P(ABC)$:

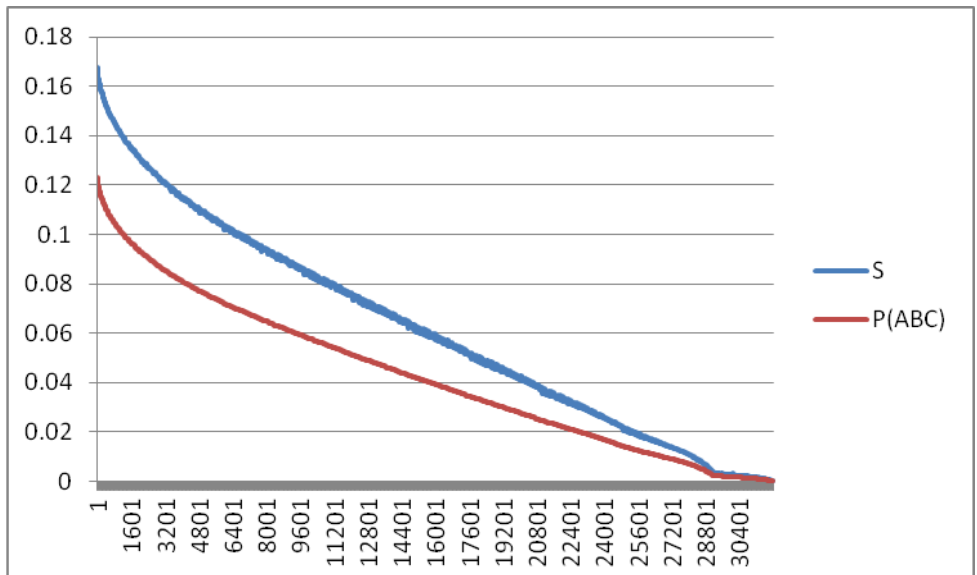

Figure 10 when the test number of is 100,000,000, the curves of the area S of the central part of Venn diagram and $P(ABC)$

We found that the change of the area S is the same as $P(ABC)$: with the $P(ABC)$ increasing, area S has a regular increase, and the curve is almost the same. Although with $P(ABC)$ increasing S have some fluctuations, in the case of test number increasing the fluctuation of S becomes smaller. In Figure 8, the curve of the area S is very broad, indicating that the data has a very large fluctuation. In Figure 9, the curve of the area S becomes narrower, indicating the volatility decreases. In Figure 10, the curve of the area S is more narrow, indicating that the amplitude of fluctuation is further reduced.

To find the true relationship between $P(ABC)$ and S, by the linear regression analysis, we conclude that:

$$P \approx kS + kS^2$$

When n = 100000000, k = 0.63, the curve of $kS + kS^2$ almost coincides exactly with the curve of $P(ABC)$ (Figure 13):

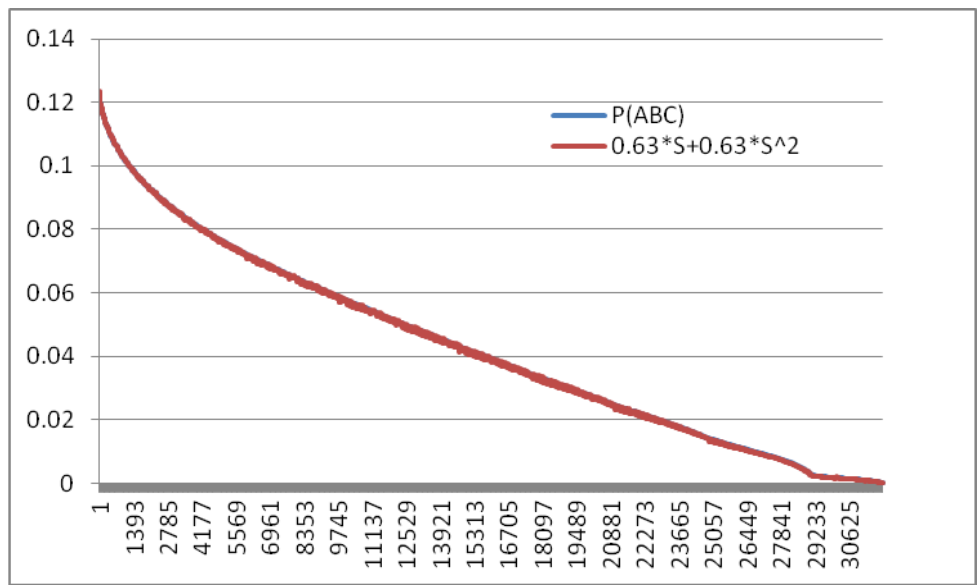

Figure 13 When k = 0.63, the curves of $kS + kS^2$ and $P(ABC)$ are almost coincide

Therefore, we give an important hypothesis that with the infinitely increase of test number, the change laws of S and $P(ABC)$ are the same, which mean the area S can represent $P(ABC)$. So, we come to an important conclusion: although S and $P(ABC)$ are not exactly the same, their sizes are positively correlated. Further, the more the number of tests, the more identical the change trend are.

Unfortunately, we are unable to prove that when the number of tests towards infinity, S and $P(ABC)$ are completely the same, and there is no fluctuation. Moreover, we cannot give a theoretical precise explanation for $kS + kS^2$ and only guess the larger the central part, the more data belongs to $P(ABC)$ and the more possibility of the simultaneous occurrence of data points.

Experiment show that, $P(ABC)$ may only depend on $P(A), P(B), P(C)$ and $P(AB), P(AC), P(BC)$. So without knowing $P(A \cup B \cup C)$, we can also get the relative size of $P(ABC)$.

The above method for three events can be extended to four events, five events and even more events. The quadruple $P(ABCD)$ can be seen as the joint probability of the three probabilities $P(AB), P(C)$ and $P(D)$. Known $P(AB), P(C), P(D)$ and $P(ABC), P(ABD), P(CD)$ and using the above method, we can figure out $P(ABCD)$. **Therefore, the overall probability of one event can be obtained by the probability of its part**, for example, the quality rate of a product depends on the quality rate of all its parts; the probability for a lottery to win depends on the correct probability of each number.